\documentclass[nohyperref]{article}

\usepackage[preprint]{neurips_2022}

\usepackage{microtype}
\usepackage{graphicx}
\usepackage{subfigure}
\usepackage{booktabs} 

\usepackage{hyperref}

\usepackage{amsmath}
\usepackage{amssymb}
\usepackage{mathtools}
\usepackage{amsthm}
\usepackage{bbm} 

\usepackage[capitalize,noabbrev]{cleveref}

\RequirePackage{times}

\RequirePackage{fancyhdr}
\RequirePackage{xcolor} 
\RequirePackage{algorithm}
\RequirePackage{algorithmic}
\RequirePackage{natbib}
\RequirePackage{eso-pic} 
\RequirePackage{forloop}
\RequirePackage{url}

\theoremstyle{plain}
\newtheorem{theorem}{Theorem}[section]

\newtheorem{lemma}[theorem]{Lemma}

\theoremstyle{definition}

\theoremstyle{remark}

\usepackage[textsize=tiny]{todonotes}

\newcommand{\A}{\mathcal{A}}
\newcommand{\X}{\mathcal{X}}
\newcommand{\R}{\mathbb{R}}
\newcommand{\Reg}{\mathcal{R}}

\newcommand{\n}{n}
\newcommand{\sumt}{\sum^{\n}_{t=1}}
\newcommand{\suma}{\sum_{a}}
\newcommand{\sumk}{\sum_{k=1}^{t-1}}
\newcommand{\One}[1]{\mathbbm{1}\{ #1 \}}

\newcommand{\iwr}{\widehat{r}} 
\newcommand{\ixr}{\widetilde{r}} 

\newcommand{\Vinv}{(\bar{V}_b)^{-1}}

\DeclarePairedDelimiterX\ip[2]{\langle}{\rangle}{#1,#2}

\let\P\undefined
\DeclarePairedDelimiterXPP\P[1]{\mathbb{P}}(){}{
    
    #1
}

\DeclarePairedDelimiterXPP\E[1]{\mathbb{E}}[]{}{
    
    #1
}

\DeclarePairedDelimiterXPP\Es[2]{\mathbb{E}_{#1}}[]{}{
    
    #2
}

\DeclareMathOperator{\Tr}{Tr}

\usepackage{xspace}

\renewcommand{\Vinv}{\pa{\V(\pib)}^{-1}}
\newcommand{\expthreeix}{\textsc{Exp3-IX}\xspace}
\newcommand{\expthree}{\textsc{Exp3}\xspace}
\newcommand{\prodd}{\textsc{Prod}\xspace}
\newcommand{\linprod}{\textsc{LinProd}\xspace}
\renewcommand{\mid}{|}
\newcommand{\pib}{\pi_B}
\newcommand{\ab}{A^B}
\newcommand{\rb}{R^B}
\newcommand{\coverage}{C}
\newcommand{\ccoverage}{C_\varphi}

\newcommand{\V}{\overline{V}}

\newcommand{\F}{\mathcal{F}}
\newcommand{\real}{\mathbb{R}}

\newcommand{\Sw}{\mathcal{S}}

\newcommand{\OO}{\mathcal{O}}

\newcommand{\trace}[1]{\mbox{tr}\left(#1\right)}

\newcommand{\EE}[1]{\mathbb{E}\left[#1\right]}

\newcommand{\PPt}[1]{\mathbb{P}_t\left[#1\right]}
\newcommand{\EEt}[1]{\mathbb{E}_t\left[#1\right]}

\newcommand{\PPcc}[2]{\mathbb{P}\left[\left.#1\right|#2\right]}

\newcommand{\EEcc}[2]{\mathbb{E}\left[\left.#1\right|#2\right]}

\def\argmax{\mathop{\mbox{ arg\,max}}}
\newcommand{\ra}{\rightarrow}

\newcommand{\siprod}[2]{\langle#1,#2\rangle}
\newcommand{\iprod}[2]{\left\langle#1,#2\right\rangle}

\newcommand{\pa}[1]{\left(#1\right)}
\newcommand{\bpa}[1]{\bigl(#1\bigr)}

\newcommand{\wh}{\widehat}
\newcommand{\wt}{\widetilde}

\newcommand{\hr}{\wh{r}}
\newcommand{\hpi}{\wh{\pi}}
\newcommand{\tr}{\wt{r}}

\newcommand{\htheta}{\wh{\theta}}

\newcommand{\transpose}{^\mathsf{\scriptscriptstyle T}}

\definecolor{PalePurp}{rgb}{0.66,0.57,0.66}

\renewcommand{\trace}[1]{\Tr{\Big( #1 \Big)}}

\title{Online Learning with Off-Policy Feedback}

\author{
	Germano Gabbianelli \\
	Universitat Pompeu Fabra\\
	Barcelona, Spain\\
	\texttt{germano.gabbianelli@upf.edu} \\
	\AND
	Matteo Papini \\
	Universitat Pompeu Fabra\\
	Barcelona, Spain\\
	\texttt{matteo.papini@upf.edu } \\
	\And
	Gergely Neu \\
	Universitat Pompeu Fabra\\
	Barcelona, Spain\\
	\texttt{gergely.neu@gmail.com} \\
}

\begin{document}

\maketitle

\vskip 0.3in

\begin{abstract}
    We study the problem of online learning in adversarial bandit problems under a partial observability model called
    off-policy feedback. In this sequential decision making problem, the learner cannot directly observe its rewards, but
    instead sees the ones obtained by another unknown policy run in parallel (behavior policy). 
    Instead of a standard exploration-exploitation dilemma, the learner has to face another challenge in this setting: 
    due to limited observations outside of their control, the learner may not be able to estimate the value of each 
    policy equally well.
    To address this issue, we propose a set of algorithms that guarantee regret bounds that scale with a natural 
    notion of mismatch between any comparator policy and the behavior policy, achieving improved performance against 
    comparators that are well-covered by the observations.
    We also provide an  extension to the setting of adversarial linear contextual bandits, and verify the theoretical 
    guarantees via a set of experiments. Our key algorithmic idea is adapting the notion of pessimistic reward 
    estimators that has been recently popular in the context of off-policy reinforcement learning.
\end{abstract}

\section{Introduction}
Off-policy learning is one of the most fundamental concepts in reinforcement learning, concerned with the problem of 
learning an optimal behavior policy given sample observations generated by a (most likely suboptimal) behavior policy. 
This setting comes with a unique set of challenges arising from the fact that the learning agent has no influence over 
the observed data, and thus classical methods for reducing uncertainty via exploration do not directly apply. The 
inability to explore may suggest that off-policy learning is better approached as a simple ``pure exploitation'' 
problem and can be potentially solved by a greedy approach---however, more thought reveals that an effective learning 
method should also attempt to account for the uncertainty of the random observations. Indeed, the problem setting comes 
with multiple layers of uncertainty: one layer being the potentially random choices made by the behavior policy, and 
another being the randomness in the observed rewards. In the present paper, we study a setting where the two 
uncertainties can be decoupled and addressed individually: the setting of online learning against an adversarial 
sequence of rewards, with off-policy feedback revealed by a stationary random policy.

The setting we study lies in the intersection of two distinct paradigms of sequential decision making: adversarial 
online learning and off-policy reinforcement learning. Concretely, we study a sequential decision making problem where 
in each round, the learner has to pick one of $K$ actions in order to maximize its total rewards. The sequence of 
reward assignments to actions are decided by an adversary, with each reward function determined the moment before the 
learner selects its action. The unique feature of the setting is that the learner does not get to observe its reward. 
However, the learner does observe the reward of another action that has been randomly sampled according to a behavior 
policy that remains fixed during the learning process. The goal of the learner is then to gain nearly as much reward as 
the best fixed comparator policy.


While there is indeed no exploration-exploitation dilemma that the learner has to address in this setting, some 
precaution in selecting the actions is still needed due to the potentially malicious choices of the adversary. 
Another, perhaps bigger, challenge is that the learner has to estimate the rewards of its own actions from observations 
made by another policy. The rewards of actions that are observed less frequently are obviously more difficult to 
estimate. It is thus a sensible requirement to have better performance guarantees against actions that have been more 
frequently played than against less well-understood actions. More generally, we aim to obtain guarantees that depend on 
how well the comparator policy is ``covered'' by the behavior policy.

Our main contribution is an online learning algorithm that guarantees a total expected regret against any comparator 
policy $\pi^*$ that is of order $\sqrt{n} \sum_a \frac{\pi^*(a)}{\pi_B(a)}$, where $\pi_B$ is the behavior policy. 
Our method makes use of a slight \emph{pessimistic} adjustment to the classic importance-weighted reward estimators 
commonly used in the adversarial bandit literature.
We refer to the problem-dependent factor appearing in the bound as the \emph{coverage ratio} and denote it by 
$\coverage(\pi^*;\pib)$. The coverage ratio quantifies the overlap between the comparator and behavior 
policies: it is of order $K$ when the two policies closely match each other, but it blows up quickly as the two 
policies start to differ. Notably, our bounds can be orders of magnitude better than what one would obtain by adapting a 
standard adversarial bandit method without adjustments. For instance, a na\"ive analysis of the classic \expthree method 
only gives a regret bound of order $\sqrt{n / \min_a \pi_B(a)}$ against all comparator policies---even against ones that 
are actually well covered by the behavior policy. Besides providing theoretical results, we also confirm 
empirically that the performance of these two methods can be quite different, and in particular that \expthree can 
indeed fail to take advantage of the comparator policy being poorly covered by the behavior policy.

Our contributions are most easily interpreted in the broader context of online learning under partial monitoring, which 
generally considers situations where the observations made by the learner are decoupled from its rewards 
\citep{Rus99,BFPRSz14,LSz19a}. In a general partial monitoring scenario, the learner receives an observation that 
depends on its action but may be insufficient to reconstruct the obtained reward. A well-studied 
special case of partial monitoring problems is online learning with feedback graphs 
\citep{MS11,ACBDK15,ACBGMMS17,KNVM14,KNV16}. In this setting, the set of observations associated with each action are 
given by a directed graph whose nodes are the actions: if actions $a$ and $a'$ are connected with an arc pointing from 
$a$ to $a'$, the learner observes the reward of action $a$ when it plays action $a'$. The graph may not have self-loops 
for every action, which allows the possibility that the learner will not observe its own reward. Clearly, our setting 
can be embedded in this class of problems by considering a sequence of randomly generated star graphs where the action 
taken by the behavior policy is connected with all other actions. However, the graph does not contain self-loops which 
renders all existing methods for this problem unsuitable for our problem. In this sense, our contribution sheds some new 
light on the hardness of learning with feedback graphs without self-loops, and can potentially inspire future work in 
this domain.

Another line of work closely related to ours is the literature on offline reinforcement learning, where the learner
cannot interact with the environment and has instead only access to a fixed dataset gathered by a behavior policy 
\citep{LKTF20}. In this context, the idea of of employing some form of pessimism has been extremely popular in the last 
few years, and pessimism has been purported to come with many desirable properties 
\citep{JinYW21,BuckmanGB21,US21,RZMJR21,XCJMA21}. One of these is that pessimistic offline RL methods can overcome the 
typical limitation of requiring the behavior policy to sufficiently explore the \emph{whole} state-action space, which 
many previous results suffer from  \citep{ASM08,MSz08,CJ19,XJ21}. This assumption is very strong and often not verified 
in practice. However, a series of recent works show that, via an appropriate use of 
pessimism, it is possible to obtain bounds which scale with the coverage with respect to a comparator policy, instead of 
the whole state-action space. Many of these results are surveyed in the work of \citet{Xiao2021}, who 
show that pessimistic policies are minimax optimal with respect to a special objective that weighs problem instances 
with a notion of inherent difficulty of estimating the value of the optimal policy. On the other hand, they show that 
without such weighting, pessimism is in fact only one of many possible heuristics that are all minimax optimal when 
considering the natural version of the optimization objective.
This highlights that pessimism may not necessarily play a 
special role in offline optimization, and that the quest to understand the complexity of offline reinforcement learning 
is far from being over. 

While our results definitely do not settle the debate of whether or not pessimism is the best way to deal with 
off-policy observations, they do provide some new insights. Most importantly, our findings highlight that pessimism 
remains an effective method for obtaining comparator-dependent guarantees. Such guarantees have attracted quite some 
interest in the literature on online learning with fully observable outcomes \cite{Chaudhuri2009, Koolen2013,
Luo2015, Koolen2015, OP16, CO18}. One common building block of parameter-free methods in this context is the \prodd 
algorithm of \citet{CBMS07}, used, for instance, in the algorithm designs of \citet{SNL14,GSvE14,Koolen2015}. 
Interestingly, our analysis also leans heavily on the tools developed by \citet{CBMS07}. When it comes to the bandit 
setting, comparator-dependent results are apparently much more sparse and in fact we are only aware of the work of 
\cite{Lattimore2015} that studies the possibility of guaranteeing better performance against certain comparators. As for 
our specific problem, we are not aware of any existing method that would be able to guarantee meaningful 
instance-dependent performance bounds.



%




\paragraph{Notation.} We denote the set of probability distributions over a set $\Sw$ as $\Delta_{\Sw}$. 
We denote the scalar product of $x,y\in\real^d$ as $\iprod{x}{y}$ and use $\|\cdot\|_2$ to denote the Euclidean norm. 
For a positive semi-definite matrix $A\in\real^{d\times d}$, we write $\lambda_{\min}(A)$
and $\Tr{(A)}$ to denote respectively its smallest eigenvalue and its trace. Finally, we 
use the conventions that $\prod_{k=i}^j = 1$ and $\sum_{k=i}^j = 0$ when $j < i$.

\section{Preliminaries}
We study an $n$-rounds sequential-decision game between a \textit{learner} (who has a finite set of actions $\A$) and an
\textit{adversary}, where the following steps are repeated in each round $t\in[\n]$:
\begin{enumerate}
    \item The adversary picks a reward function $r_t:\A\to[0,1]$, mapping each
        action to a nonnegative numerical reward,
    \item an action $A_t\in\A$ is selected by the learner, 
    \item another action $\ab_t\in \A$ is selected according to a fixed \emph{behavior policy} $\pib$,
    \item the learner gains reward $R_t = r_t(A_t)$ and observes $\rb_t = r_t(\ab_t)$ along with $\ab_t$.
\end{enumerate}
Notably, the learner does not get to observe its own reward $R_t$ but has to make do with the reward $\rb_t$ gained by 
the behavior policy. We allow the adversary to be adaptive in the sense of being able to take into account all past 
actions of the learner and the behavior policy when selecting the reward function. Also, the learner is 
allowed to use randomization for selecting its action. Precisely, we will denote the history of interactions up to the 
end of round $t$ by $\F_t = \sigma\pa{r_t,\ab_t,A_t,\dots,r_1,\ab_1,A_1}$, and respectively denote conditional 
expectations and probabilities with respect to the induced filtration by $\EEt{\cdot} = \EEcc{\cdot}{\F_{t-1}}$ and 
$\PPt{\cdot} = \PPcc{\cdot}{\F_{t-1}}$. With this notation, we define the \emph{policy} of the learner as $\pi_t(a) = 
\PPt{A_t = a}$ for all actions $a\in\A$.

The objective of the learner is to minimize the \emph{regret}, with respect
to any time-invariant comparator policy $\pi^*\in\Delta_\A$, defined as
\begin{equation}
\Reg(\pi^*) = \sum^\n_{t=1} \sum_a  \bpa{\pi^*(a)-\pi_t(a)}r_t(a).
\end{equation}
In particular we will be interested in providing bounds on $\E{\Reg(\pi^*)}$, where
the expectation integrates over all the randomness involved in selecting the random actions $A_t$ and $\ab_t$ and 
the reward functions $r_t$. In words, the expected regret measures the expected gap between the total rewards gained by 
the learner and the amount gained by a fixed comparator policy $\pi^*$.

The most common definition of regret compares the learner's performance to the optimal policy $\pi^*$ that selects the 
action $a^* = \argmax_a \sum_{t=1}^\n r_t(a)$. However, it is easy to see that this comparator strategy may be 
unsuitable for measuring performance in the setting we consider. Specifically, it is unreasonable to expect strong 
guarantees against the optimal policy when the behavior policy selects the optimal actions very rarely. Specifically, 
the adversary can take advantage of the behavior policy covering the action space only partially, and hide the best 
rewards among the least-frequently sampled actions. In the most extreme case, the behavior policy may not select some 
actions at all, which clearly makes it impossible for the learner to compete with the optimal policy. Thus, we aim to 
achieve regret guarantees that scale with the level of mismatch between the behavior and comparator policies, capturing 
the intuition that comparator strategies that are well covered by the data should be easier to compete with. 
Concretely, we will define the coverage ratio between $\pi^*$ and $\pib$ as 
\begin{equation}\label{eq:coverage_def}
 \coverage(\pi^*;\pib) = \sum_a \frac{\pi^*(a)}{\pib(a)},
\end{equation}
and aim to provide regret bounds that scale with this quantity.
The intuitive significance of this coverage ratio is that it roughly captures the hardness of estimating the value of 
the comparator policy $\pi^*$ using only data from $\pib$. Indeed, a simple argument reveals that the estimation error 
of the total reward of any given action $a$ scales as $\sqrt{n/\pi_B(a)}$ in the worst case. Thus, we set out to prove 
regret guarantees against each comparator $\pi^*$ that scale proportionally to the worst-case estimation error of order 
$\sqrt{C(\pi^*;\pib) n}$.

\section{Algorithm and main results}
This section presents our main contributions: a set of algorithms for online off-policy learning and their
comparator-dependent performance guarantees that scale with the coverage ratio between the comparator policy and the
behavior policy. For the sake of clarity of exposition, we first describe our approach in a relatively simple setting
where the number of actions is finite and the behavior policy is known. We then extend the algorithm to be able to deal
with unknown behavior policies in Section~\ref{sec:unknown_pi} and to linear contextual bandit problems in
Section~\ref{sec:linear}.

\begin{algorithm}
	\caption{\expthreeix for Off-Policy Learning}\label{alg:exp3ix}
	\begin{algorithmic}
        \STATE {\bfseries Input:} learning rate $\eta$, IX parameters $\pa{\gamma_t}_{t=1}^n$
		\FOR{$t\gets 1,\dots,n$}
        \STATE compute $w_t(a) = \exp(\eta \sumk\ixr_k(a))\quad\forall a\in\A$
		\STATE play $A_t$ according to $\pi_t(\cdot) = w_t(\cdot)/ \sum_{a\in\A} w_t(a)$
		\STATE observe $\ab_t$ and $\rb_t$
		\STATE compute $\ixr_t(\ab_t) = \rb_t /\pa{\pib(\ab_t)+\gamma_t}$
		\ENDFOR
	\end{algorithmic}
\end{algorithm}

\subsection{Known behavior policy}
Let us first consider the case where the learner has full prior knowledge of $\pib$. The algorithm we propose is an
adaptation of the \expthreeix algorithm first proposed by \citet{KNVM14} and later analyzed more generally by
\cite{Neu2015}. At each time-step $t$ the algorithm computes the weights
\begin{align*}
    w_1(a) &= 1,\\
    w_t(a) &= w_{t-1}(a)e^{\eta\, \ixr_{t-1}(a)} = \exp\pa{\eta\sumk \ixr_k(a)},
\end{align*}
and the normalization factors $W_{t} = \suma w_t(a)$, and uses them to draw the action $A_t$ according to $\pi_t(a) =
	\frac{w_t(a)}{W_t}$. Here, $\eta$ is a positive learning-rate parameter and $\ixr$ is the \emph{Implicit
	eXploration} (IX) estimate
of the reward function $r_t$, modified to use the rewards obtained by the behavior policy $\pib$, since the learner
cannot see its own rewards:
\begin{equation}\label{eq:rhat}
	\ixr_{t}(a) = \frac{\rb_t \One{\ab_t = a }}{\pib(a) + \gamma_t} =
	\frac{r_{t}(a) \One{\ab_t = a }}{\pib(a) + \gamma_t},
\end{equation}
where $\gamma_t\geq 0$ is an appropriately chosen parameter. The full algorithm is shown as \Cref{alg:exp3ix} .

When setting $\gamma_t = 0$, $\ixr_t$ is clearly an unbiased estimator of $r_t$ since $\EEt{\One{\ab_t = a}} =
	\pib(a)$. Otherwise, for $\gamma_t > 0$, the estimator is biased towards zero which can be seen as a
\emph{pessimistic} bias in the sense that it underestimates the true rewards: $\EEt{\ixr_t(a)} \le r_t(a)$. As our
results below show, this property is crucially important to achieve our goal to obtain performance guarantees that scale
with the mismatch between $\pi^*$ and $\pib$ \footnote{Notably, the pessimistic property
	of the IX estimator also relies on the fact that we consider positive rewards. When working with negative rewards (or
	losses), the estimator is optimistically biased which is exploited by the high-probability analysis of
	\citet{Neu2015}.} . In particular, our main result regarding this method is the following:\looseness=-1
\begin{theorem}\label{thm:known_pi}
	For any comparator policy $\pi^*$, the expected regret of \expthreeix initialized with any positive
	learning rate $\eta$ and $\gamma_t = \frac{\eta}{2}$, is bounded as
	\begin{equation}\label{eq:bound_full}
		\E{\Reg(\pi^*)} \le \frac{\log K}{\eta} + \frac{\eta}{2} \sumt \suma	\frac{r_t(a)
			\pi^*(a)}{\pi_B(a) + \frac{\eta}{2}}.
	\end{equation}
	Setting the learning rate to $\eta = \sqrt{\frac{\log K}{n}}$ and to $\eta = \sqrt{\frac{\log
				K}{\coverage(\pi^*;\pib) n}}$ respectively gives
	\begin{align}\label{eq:bound_uniform}
        \E{\Reg(\pi^*)} &\le
		\sqrt{n\log{K}}\pa{1 + \frac 12 \coverage(\pi^*;\pib)} \\
        \E{\Reg(\pi^*)}&\le\sqrt{2\coverage(\pi^*;\pib) n\log{K}}.
	\end{align}
\end{theorem}
The proof is based on a set of small but important changes made to the standard \expthree analysis originally due to
\citet{ACBFS02}, and is deferred to Section~\ref{sec:analysis}.
The bound above successfully achieves our goal of guaranteeing better regret against comparator policies that are
well-covered by the behavior policy. In particular, the first bound of Equation~\eqref{eq:bound_uniform} provides a
bound that holds uniformly for all behavior policies without requiring prior commitment to any coverage level, whereas
the second bound guarantees improved guarantees against policies with a given coverage level at the
price of using a learning-rate parameter that is specific to the desired coverage. Notably, the coverage ratio is of
the order $K$ when the comparator policy closely matches the behavior policy, but the actual bound of
Equation~\eqref{eq:bound_full} can be much smaller when there are many actions that the behavior policy selects with
probability much smaller than $\gamma$.

It is worthwhile to compare this result with what one would obtain by a straightforward adaptation of a standard
adversarial bandit algorithm like \expthree \citep{ACBFS02}---which essentially corresponds to our algorithm with the
choice $\gamma = 0$. A standard calculation shows that the regret of this strategy can be upper bounded by
\[
	\E{\Reg(\pi^*)} \le \frac{\log K}{\eta} + \eta \sumt \suma \EE{\frac{\pi_t(a)}{\pi_B(a)}}.
\]
Notice that the right-hand side of this bound does not depend on the comparator policy, which suggests that this method
is not quite suitable for achieving our goal. Even worse, the only way to bound the second term in the bound seems to
be by $n / \min_a \pib(a)$, which scales inversely with the coverage of the least well-covered action. A pessimistic
interpretation of this argument suggests that \expthree may have huge regret when some actions are not covered
appropriately. A more charitable reading is that \expthree may not be able to take advantage of situations where the
comparator policy is well-covered by the behavior policy. We set out to understand this phenomenon
empirically in Section~\ref{sec:experiments}.\looseness=-1

\subsection{Unknown behavior policy}\label{sec:unknown_pi}
In the previous section we assumed to have full prior knowledge of the behavior policy $\pib$ in order to compute our
reward estimator $\iwr_t$. In this section, we show that this is not an inherent limitation of our technique and
that it can be easily addressed by using a simple plugin estimator $\wh{\pi}_t$ of the behavior policy, which is then
used in the definition of $\ixr_t$:
\begin{align}\label{eq:pit}
    \hpi_1(\cdot) &= 0, \qquad\qquad\hpi_t(a) = \frac{1}{t-1}\sum_{k=1}^{t-1} \One{\ab_k = a}, \\
    \ixr_{t}(a) &= \frac{r_{t}(a) \One{\ab_t = a }}{\hpi_t(a) + \gamma_t}.
\end{align}
We then feed these reward estimates to the exponential-weights procedure described in the previous section. As
the following theorem shows, the resulting algorithm satisfies essentially the same regret bound as the method
that has full knowledge of $\pib$.
\begin{theorem}\label{thm:unknown_pi}
	For any comparator policy $\pi^*$, the expected regret of \expthreeix with
	learning rate $\eta=\sqrt{\log(K)/\n}$ and parameter sequence $\gamma_1
		= 1 + \frac{\eta}{2}$, $\gamma_t = \frac{\eta}{2} + \sqrt{\log(K(t-1)^2)/(2t-2)}$,
	and estimates as in Equation~\eqref{eq:pit}, is bounded as
	\begin{equation}\label{eq:bound_unknown}
		\E*{\Reg(\pi^*)} = \OO\pa{C(\pi^*;\pib)\sqrt{n\log(Kn)}}.
	\end{equation}
\end{theorem}
The parameter tuning achieving the above bound is similar to what is used in the previous theorem, and does not require
the learner to have any problem-specific information that would be difficult to acquire. Details are relegated to
Appendix~\ref{app:unknown_pi} along with the proof of the theorem.

\subsection{Linear contextual bandits}\label{sec:linear}
We now switch gears and provide an extension to a significantly more advanced setup: that of adversarial linear
contextual bandits, first studied by \citet{Neu2020}. In each round $t$ of this sequential game, the learner first
observes a context $X_t$ before making its decision, and the reward function $r_t$ is assumed to be an adversarially
chosen function of the context $X_t$ and the action $A_t$ taken by the learner. In particular, the adversary chooses
a \emph{reward vector} $\theta_t\in\R^d$ at each step, which determines the rewards for each context-action pair as
$r_t(x,a) = \iprod{\theta_t}{\varphi(x,a)}$, where $\varphi:\X\times\A\to\R^d$ is a \textit{feature map}
known to both the learner and the adversary. We assume that the contexts live in an abstract space $\X$ and are 
drawn i.i.d.~according to a fixed probability distribution for all $t$. On the other hand, the 
adversary has full freedom in choosing the reward functions, as long as it only depends on past observations and in 
particular does not depend on $X_t$ or $A_t$. The only restriction we put on the adversary is that we continue to 
require the rewards to be in the interval $[0,1]$.
Moreover, as in the previous section the learner is not allowed to see its own rewards, but only the ones
of an other policy $\pib$ running in parallel. In this setting, a policy $\pi$ is a mapping from contexts to
probability distributions over the space of actions.


The objective of the learner is to minimize the regret defined with respect
to any time-invariant comparator policy $\pi^*:\X\ra\Delta_{\A}$ as:
\[
	\Reg(\pi^*) = \sumt \suma \left(\pi^*(a \mid X_t) - \pi_t(a \mid X_t)\right) r_t(X_t, a).
\]
%
Our algorithm for this setting is a combination of the context-wise exponential weights method proposed by
\citet{Neu2020} with the ideas developed in the previous section. The algorithm design is complicated by the fact that
the implicit exploration estimator is not very straightforward to extend to this setting, which necessitates an 
alternative (but closely related) approach. In particular, we will define an \emph{unbiased} estimator of the reward 
vector $\theta_t$
and feed the resulting reward estimates to calculate policy updates via the \prodd update rule proposed by
\citet{CBMS07} (see also \citet{CesaBianchi2006}, Section 2.7).

Concretely, following the algorithm design of \citet{Neu2020}, we define the matrix $\V(\pi) =
	\EEt{\sum_a \pi(a|X_t) \varphi(X_t,a)\varphi(X_t,a)\transpose}$ and the estimator
\begin{equation}\label{eq:theta_hat}
	\wh{\theta}_t = \pa{\V(\pi_B)}^{-1} \varphi(X_t,\ab_t)\rb_t.
\end{equation}
Since $\varphi(X_t,\ab_t)\rb_t = \varphi(X_t,\ab_t)\varphi(X_t,\ab_t)\transpose \theta_t$, it is easy to see that
$\htheta_t$ is an unbiased estimator of $\theta_t$. These estimators are then used to update a set of weights
$w_t$ defined for each context-action pair as\looseness=-1
\begin{align*}
	&w_t(x,a) = \prod_{k=1}^{t-1} (1+\eta\siprod{\htheta_k}{\varphi(x,a)}),\\&  W_t(x) = \sum_a w_t(x,a),
\end{align*}
and the policy is then given as $\pi_t(a|x) = \frac{w_t(x,a)}{W_t(x)}$. Notice that the policy can be easily
implemented without explicitly keeping track of the weights for all $(x,a)$ pairs, as they are well-defined through the
sequence of reward-estimate vectors $\wh{\theta}_{1},\dots,\wh{\theta}_{t-1}$.
For simplicity\footnote{These restrictions can be removed using the techniques developed in the previous section, 
although at the price of a significantly more technical analysis. We opted to preserve clarity of presentation 
instead.}, we assume that the matrix $\V(\pi_B)$ is known for the learner and has uniformly lower-bounded eigenvalues 
so that its inverse exists. 
Following the naming convention of \citet{Neu2020}, we call the resulting algorithm \linprod and show its pseudocode in 
Algorithm~\ref{alg:prod-ctx}.

%

\begin{algorithm}
	\caption{\linprod for off-policy learning}\label{alg:prod-ctx}
	\begin{algorithmic}
        \STATE {\bfseries Input:} learning rate $\eta$
		\FOR{$t\gets 1,\dots,n$}
		\STATE observe $X_t$
		\STATE compute $w_t(X_t, \cdot) = \prod_{k=1}^{t-1}
			\bpa{1+\eta\siprod{\htheta_k}{\varphi(X_t,\cdot)}}$
		\STATE draw $A_t$ from $\pi_t(\cdot\mid X_t) = w_t(X_t,\cdot)/\sum_a w_t(X_t,a)$
		\STATE observe $\rb_t$ and $\varphi(X_t,\ab_t)$
        \STATE compute $\wh{\theta}_{t}$ as in \cref{eq:theta_hat}
		\ENDFOR
	\end{algorithmic}
\end{algorithm}

%
%
%
%
%
%
%
%

Similarly to the previous sections, we are aiming for a comparator-dependent performance guarantee that depends on the
mismatch of the comparator and the behavior policy. However, this quantity is not straightforward to define in the case
that we consider, due to the fact that we consider a potentially infinite space of contexts. In particular, the natural
idea of considering $\EE{\sum_a \frac{\pi^*(a|X_t)}{\pib(a|X_t)}}$ as a measure of mismatch is problematic as it can
blow up when there exists even a tiny set of contexts where the two policies have no overlap. Intuitively, it should be
possible to estimate the reward vector even when there are states where the policies pick different actions, as long as
they are aligned in the feature space in an appropriate sense. To make this intuition formal, we will consider the
following alternative notion of \emph{feature coverage ratio}:
\begin{equation}\label{eq:ccoverage}
\ccoverage(\pi^*;\pib) = \Tr\Big[ {\pa{\V(\pib)}^{-1}\, \V(\pi^*)} \Big].
\end{equation}
This notion of coverage appropriately measures the extent to which the feature vectors $\varphi(X_t,A_t^*)$ generated
by the comparator policy line up with the features excited by the behavior policy. Similar distribution-mismatch 
measures are common in the offline RL literature, and in particular the results of \citet{JinYW21} are stated in terms 
of the same quantity.
The following theorem gives a
performance guarantee stated in terms of this measure of distribution mismatch.\looseness=-1
\begin{theorem}\label{thm:linear}
	Let $\eta$ be any positive learning rate and suppose that it is small enough so that 
	$\lambda_{\min}(\V(\pib)) \geq 2\eta\,\sup_{x,a} \left\|\phi(x,a)\right\|^2_2$ holds.
	Then, for any comparator policy $\pi^*$ the expected regret of \linprod is upper-bounded by
	$$
		\E[\big]{\Reg(\pi^*)} \leq \frac{\log K}{\eta} + \eta n \ccoverage(\pi^*;\pib),
	$$
Setting $\eta=\sqrt{\frac{\log K}{n}}$ and $\eta=\sqrt{\frac{\log K }{\ccoverage(\pi^*;\pib)n}}$ and supposing that $n$ 
is large enough so that $\eta$ satisfies the condition, the regret can be further bounded respectively as
	\begin{align*}
        \E{\Reg(\pi^*)} &\leq \sqrt{n\log K}\left(
		1 + \ccoverage(\pi^*;\pib)
		\right),\\
        \E{\Reg(\pi^*)} &\leq 2\sqrt{\ccoverage(\pi^*;\pib) n\log K}.
	\end{align*}
\end{theorem}
The bound mirrors the qualities of Theorem~\ref{thm:known_pi}, and in particular it implies good performance when the
comparator policy is well-covered by the behavior policy. Under ideal conditions where these policies are close enough,
the coverage ratio is of order $d$, which essentially matches the rate proved by \citet{Neu2020} for the case of
standard bandit feedback. The bound then degrades as the two policies drift apart.

%
%

\section{Analysis}\label{sec:analysis}
This section provides the key ideas required for proving our main results. Due to space restrictions, we will only
prove Theorem~\ref{thm:known_pi} here and defer the proof of the other two theorems to Appendices~\ref{app:unknown_pi}
and~\ref{app:linear}.\looseness=-1
%
%

For the analysis, it will be useful to define the unbiased reward estimator
$\hr_t(a) = \frac{r_t(a) \One{\ab_t = a}}{\pib(a)}$,
which essentially corresponds to the biased IX estimator $\tr_t$ when setting $\gamma_t = 0$.
One of the most important properties of the IX estimator that we will repeatedly use is stated in the following
inequality:
\begin{equation}\label{eq:IX_vs_prod}
	\frac{r_t(a) \One{\ab_t = a}}{\pib(a) + \gamma_t} \leq \frac{1}{2\gamma_t}\log\pa{1 + 2\gamma_t \iwr_t(a)}.
\end{equation}
The result follows from a simple calculation in the proof of Lemma~1 of \citet{Neu2015} that we reproduce in
Appendix~\ref{app:IXcalc} for the convenience of the reader. Notably, the term
on the right hand side can be thought of as a reward estimator itself. Combining this reward
estimator with the exponential weights policy with $\eta = \gamma$ gives rise to the \prodd algorithm of
\citet{CBMS07}, which is a fact that some of our proofs will implicitly take advantage of. This observation also
motivates our algorithm design for the contextual bandit setting in Section~\ref{sec:linear}.

%
%
%
%
%

\paragraph{The proof of Theorem~\ref{thm:known_pi}}
The proof builds on the classical analysis of exponential weights algorithm originally due to \citet{Vov90},
\citet{LW94} and \citet{FS97}, and its extension to adversarial bandit problems by \citet{ACBFS02}. In particular, our
starting point is the following lemma that can be proved directly with arguments borrowed from any of these past works:
\begin{lemma}\label{lem:rbound_partial}
    \begin{align*}
        &\sumt \suma \pi^*(a) \tr_t(a) 
        \le \frac{\log K}{\eta} \\&\qquad+ \frac{1}{\eta}\sumt \log \suma \pi_{t}(a)
		\exp\pa{\eta \tr_{t}(a)}.
    \end{align*}
\end{lemma}
We include the proof for the sake of completeness in Appendix~\ref{app:rbound_partial}.
To proceed, notice that the above bound can be combined with Equation~\eqref{eq:IX_vs_prod} to obtain
\begin{equation}\label{eq:almostbound}
    \begin{aligned}[b]
		&\sumt \suma \pi^*(a) \tr_t(a)                                                            \leq  \frac{\log K}{\eta}  \\
            &\qquad+\frac{1}{\eta}\sumt \log \suma \pi_{t}(a)
		\exp\pa{\frac{\eta}{2\gamma}\log\pa{1+ 2\gamma\,\iwr_t(a)}}\\
		&\quad= \frac{\log K}{\eta} + \frac{1}{\eta}\sumt \log \suma \pi_{t}(a) \pa{1+ \eta \iwr_t(a)} \\
		&\quad\le \frac{\log K}{\eta} + \sumt \suma \pi_{t}(a) \iwr_t(a),
	\end{aligned}
\end{equation}
where we used the choice $\gamma = \eta/2$ in the second line and the inequality $\log(1+x) \le x$ that
holds for all $x>-1$ in the last line.


It remains to relate the two sums in the above expression to the total reward of the learner and the comparator policy.
To this end, we first notice that for any given action $a$, we have
\begin{equation}\label{eq:ixr_to_r}
	\begin{aligned}
		\EEt{\ixr_{t}(a)}
		 & = \EEt{\frac{r_t(a) \One{\ab_t = a}}{\pib(a) + \gamma}} \\
		 & = \frac{r_t(a) \pib(a)}{\pib(a) + \gamma}
		= r_t(a) - \frac{\gamma r_{t}(a)}{\pib(a) + \gamma}.
	\end{aligned}
\end{equation}
Via the tower rule of expectation, this implies
\begin{align*}
	 & \EE{\sumt \suma \pi^*(a) \tr_t(a)} \\&= \EE{\sumt \suma \pi^*(a) r_t(a)} - \gamma\sumt \suma \frac{\pi^*(a)r_{t}(a)}{\pib(a) +
	\gamma}.
\end{align*}
Similarly, since $\EEt{\hr_t(a)} = r_t(a)$, we also have
\[
	\EE{\sumt \suma \pi_t(a) \hr_t(a)} = \EE{\sumt \suma \pi_t(a) r_t(a)}.
\]
Putting these two facts together with Equation~\eqref{eq:almostbound}, we obtain the result claimed in the theorem.

\section{Empirical Results}\label{sec:experiments}
\begin{figure}[t]
	\vskip 0.2in
	\begin{center}
		\centerline{\includegraphics[width=\columnwidth]{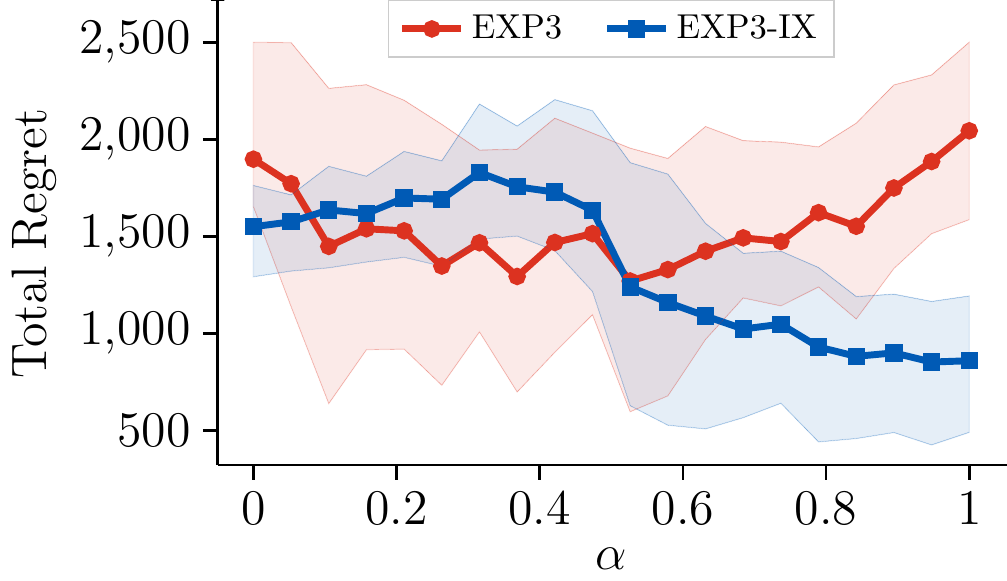}}
		\caption{Total regret after $1000$ steps for different values of the interpolation parameter $\alpha$. Thick lines
			represent the mean regret over $100$ independent runs, while the shaded area represents the interval between the $25\%$
			and $75\%$ quantiles.}
		\label{fig:experiment}
	\end{center}
	\vskip -0.2in
\end{figure}
This section provides a set of experiments to illustrate our main results, and in particular to verify if \expthreeix
can indeed take advantage of behavior policies that are well-aligned with the comparator. In particular,
we compare the performance of \expthreeix (\Cref{alg:exp3ix}) and \expthree in the
the following experiment.
We instantiate a $100$-armed bandit, with Bernoulli rewards for all arms. By default, all rewards have mean $0.5$.
However, for the first half of the game ($t\leq \n/2$), we change the mean
reward of the last arm to $0.8$, and for the remaining half, the mean
of the first arm to $1$. This means that arm $100$ is the best for the first half
of the game, but eventually gets outperformed by arm $1$.
We set the number of rounds $\n$ to $10 000$, the learning rate $\eta$ of both
algorithms to the recommended $\sqrt{\log(K)/\n}$ and $\gamma_t= \eta/2$.

We repeat the game for a range of behavior policies defined for each $\alpha$ as
$\pi_{B,\alpha}(i) \propto (1-\alpha)\frac{i}{K} + \alpha(1 - \frac{i - 1}{K})$, for
$i\in[0,\dots,K]$, where $\alpha$ varies from $0$ to $1$. Hence, $\alpha$ closer to $1$ means the behavior policy
puts large probability mass on the first action, which we use as the comparator in our experiment.
We plot the results of the experiment on Figure~\ref{fig:experiment}
The results clearly match the intuitions that one can derive from our performance guarantees. First, recall that a
na\"ive analysis of \expthree suggests that its regret may scale as $\sqrt{\n/\min_a \pib(a)}$ in the worst case.
Our empirical results show that this crude analysis is not entirely off track: the regret of \expthree indeed
deteriorates as $\min_a \pib(a)$ approaches $0$ at the two extremes $\alpha = 0$ and $\alpha = 1$. In particular,
\expthree fails to take advantage of the favorable case where the optimal policy is well-covered and performs
similarly as in the case where the optimal policy is not covered at all. In comparison, \expthreeix performs
significantly better when the comparator policy is well covered, as predicted by our theory. Other experiments on
the same problem with alternative parametrizations gave qualitatively similar results.




\section{Conclusion}
We introduced a new online learning setting where the learner is only allowed to observe off-policy feedback generated 
by a fixed behavior policy. We have proposed an algorithm with comparator-dependent regret bounds of order 
$C(\pi^*;\pib)\sqrt{n}$, depending on a naturally defined coverage ratio parameter $\coverage(\pi^*;\pib)$ that 
characterizes the mismatch between the behavior and the comparator policies. Many questions remain open regarding the 
potential tightness of this result. First, we have shown that the bounds can be improved to $O(\sqrt{C(\pi^*;\pib) 
n})$, if one wishes to restrict their attention to comparators whose coverage level is at a fixed level 
$C(\pi^*;\pib)$. However, the tuning required for achieving this result depends on the desired coverage level. It is an 
interesting open problem to find out if this requirement can be relaxed, and  bounds of order $\sqrt{C(\pi^*;\pib) n}$ 
can be simultaneously achieved for all comparators $\pi^*$ by a single algorithm. We conjecture that this question can 
be addressed by a careful adaptation of existing techniques for adaptive online learning, and in particular we believe 
that adapting the methodology of \citet{Koolen2015} should be especially suitable for achieving this goal.

Questions regarding the best achievable performances for our newly defined problem are even more exciting. As an 
adaptation of the results of \citet{Xiao2021} show via an online-to-batch reduction, the minimax regret of any 
algorithm for this setting has to scale as $\sqrt{n/\min_a\pib(a)}$, suggesting that our na\"ive adaptation of \expthree 
is already minimax optimal. In our view, this makes it all the more interesting to identify characteristics of 
individual problem instances that make faster learning possible, and we believe that comparator-dependent regret bounds 
scaling with the coverage ratio are only one of many possible flavors of adaptive performance guarantees. One concrete 
question that we are particularly interested in is a better understanding of the ``Pareto regret frontier'' of 
achievable regrets, roughly corresponding to the set of comparator-dependent regret bounds that are achievable by any 
algorithm. Clearly, the bounds we achieve are just singular elements of this set. We conjecture that bounds of order 
$\sqrt{C(\pi^*;\pib) n}$ are indeed on the regret frontier. Whether this is indeed true or if there are other 
distinguished entries on the Pareto frontier with desirable properties remains to be seen. All in all, our results 
highlight that off-policy learning is a field of study that's ripe with open questions that can be interesting for the 
online learning community that is typically very keen on instance-dependent analysis.

A more ambitious question for future research is if our techniques can be extended to more challenging settings, and 
especially online learning in Markov decision processes \citep{even-dar09OnlineMDP,neu10o-ssp,neu14o-mdp-full}. We 
think that an extension to this setting would be particularly valuable, given the recent flurry of interest in offline 
reinforcement learning. In this context, we could potentially exploit the unique feature of our algorithm design that, 
unlike all other methods, it does not rely on explicit uncertainty quantification for calculating its pessimistic 
updates. This could mean a major advantage over traditional off-policy RL methods that rely on uncertainty 
quantification to build confidence sets over abstract objects (like the entire transition function of the Markov 
process), which is a notoriously hard problem, especially in the infinite-horizon setting.
In contrast, as our results in Section~\ref{sec:linear} highlight, the pessimistic nature of our method is realized 
through an update rule that is slightly more conservative than the standard exponential-weights update rule. We believe 
that this insight can be very useful for developing new methods for offline RL, even more so since they appear to be 
directly compatible with the primal-dual off-policy learning methods of \citet{NCDL19,NDKCLS19,UHJ20}.




\bibliography{references/references,references/shortconfs}
\bibliographystyle{plainnat}

\newpage

\appendix
\onecolumn

\section{Omitted details from the proof of Theorem~\ref{thm:known_pi}}

\subsection{Proof of the bound of Equation~\eqref{eq:IX_vs_prod}}\label{app:IXcalc}
This proof is extracted from the proof of Lemma~1 of \cite{Neu2015}.
Let $c\in\R_+$ be any non-negative constant. Then,
\[
    \frac{r_t(a) \One{A_t = a}}{\pi_B(a) + c}
    \leq \frac{r_t(a) \One{A_t = a}}{\pi_B(a) + c\,r_t(a)}
    = \frac{\One{A_t = a}}{2c}\cdot\frac{{2c\,r_t(a)}/{\pi_B(a)} }{1 + {c\,r_t(a)}/{\pi_B(a)}}
    \leq \frac{1}{2c}\log( 1+2c\,\iwr_t(a) )
\]
where the first step follows from $r_t(a)\in[0,1]$ and the last one from
the inequality $\frac{x}{1+x/2} \leq \log (1+x)$, which holds for all $x\geq 0$.

\subsection{The proof of Lemma~\ref{lem:rbound_partial}}\label{app:rbound_partial}
We study the evolution of the potential function $\frac{1}{\eta} \log \frac{W_{\n+1}}{W_1}$. On the one hand, we have 
for any action $\bar{a}$ that
	\begin{equation}
    \frac{1}{\eta} \log \frac{W_{\n+1}}{W_1} = \frac{1}{\eta}\log \pa{\frac{1}{K}\suma w_{\n+1}(a)}
		\geq  \frac{1}{\eta} \log \pa{\frac{1}{K}\, w_{\n+1}(\bar{a})}
		= \sumt \ixr_t(a) - \frac{\log K}{\eta}.
	\end{equation}
	Multiplying this bound with $\pi^*(\bar{a})$ and summing up over actions gives
	the lower bound
	\begin{equation}\label{eq:exp3lower}
	\frac{1}{\eta}\log \frac{W_{\n+1}}{W_1} \ge \sumt \suma \pi^*(a)\ixr_t(a) - \frac{1}{\eta} \log K.
	\end{equation}
	On the other hand, the potential can be rewritten as follows:
	\begin{align*}
		\frac{1}{\eta}\log \frac{W_{\n+1}}{W_1}
		 & = \frac{1}{\eta}\sumt \log \frac{W_{t+1}}{W_t}
		= \frac{1}{\eta} \sumt \log \frac{\suma w_{t}(a) e^{\eta \ixr_t(a)}}{W_t}
		= \frac{1}{\eta} \sumt \log \suma \pi_{t}(a) e^{\eta \ixr_t(a)}.
	\end{align*}
	Putting the two expressions together concludes the proof.

\section{The proof of Theorem~\ref{thm:unknown_pi}}\label{app:unknown_pi}
In this proof, we have to face the added challenge of having to account for the possible inaccuracy of our estimator of
$\pib$. To this end, we define a sequence of ``good events'' under which the policy estimate is well-concentrated and
analyze the regret under this event and its complement, using that the good event should hold with high probability.
Concretely, we define the failure probability $\delta_t\in(0,1)$, the tolerance
parameter $\varepsilon_t$, and the $t$-th good event as follows:
\begin{align}
    \varepsilon_1 = 1\,,\quad\varepsilon_t = \sqrt{\frac{\log(K/\delta_t)}{2(t -1)}} && E_t = \big\{ |\hpi_t(a) - 
\pib(a)|\le \varepsilon_t\,\,(\forall a\in\A) \big\}.
\end{align}
An application of Hoeffding's inequality shows that $E_t$ holds with probability at least $1-\delta_t$.
Now, setting $\gamma_t = \varepsilon_t + \eta/2$, we can observe that under event $E_{t}$, we have
\begin{equation}\label{eq:hpbound}
	\ixr_t(a) = \frac{r_t(a)\One{\ab_t = a}}{\hpi_{t} + \gamma_t}
	\leq \frac{r_t(a)\One{\ab_t = a}}{\pib(a) + \eta/2} \leq \frac{1}{\eta}\log(1+\eta \iwr_t(a)).
\end{equation}
We proceed by noticing that the bound of Lemma~\ref{lem:rbound_partial} continues to apply, and that we can bound the
term appearing on the right-hand side as follows:
\begin{align*}
	 & \EEt{\frac{1}{\eta}\log \suma \pi_{t}(a) \exp\pa{\eta \tr_{t}(a)}} \le \One{E_{t}} \EEt{\frac{1}{\eta} \log \suma
		\pi_{t}(a) \exp\pa{\eta \tr_{t}(a)}} + \One{\overline{E}_{t}} \frac{2}{\eta}
	\\
	 & \qquad\qquad \le \One{E_{t}}\EEt{\suma \pi_{t}(a) \hr_t(a)} + \One{\overline{E}_{t}} \frac{2}{\eta} \le
	\suma \pi_{t}(a) r_t(a) + \One{\overline{E}_{t}} \frac{2}{\eta},
\end{align*}
where in the first line we used that $e^{\eta \tr_t(a)} \le e^{\eta / \gamma_t} \le e^2$
and in the second line
we used the bound of Equation~\eqref{eq:hpbound}, the fact that $E_{t}$ is $\F_{t-1}$-measurable, that $\EEt{\hr_t(a)}
	= r_t(a)$, and finally upper bounded the indicator $\One{E_{t}}$ by one.
Taking marginal expectations and summing up for all $t$, we get
\[
	\EE{\frac{1}{\eta}\sum_{t=1}^n \log \suma \pi_{t}(a) \exp\pa{\eta \tr_{t}(a)}} \le \EE{\sumt \suma \pi_{t}(a) 
r_t(a)}
	+ \frac{2}{\eta}\sumt \delta_t,
\]
where we used $\EE{\One{\overline{E}_t}} \le \delta_t$.

It thus remains to relate the term on the left-hand side of the bound of Lemma~\ref{lem:rbound_partial}. To do this, we
similarly write
\begin{align*}
	\EEt{\ixr_t(a)} & \ge \One{E_{t}} \EEt{\ixr_t(a)} =
	\One{E_{t}}\EEt{\frac{r_t(a)\One{\ab_t = a}}{\hpi_{t}(a) + \gamma_t}}
	\\
	                & \geq \One{E_{t}}\cdot \frac{r_t(a)\pib(a)}{\pib(a)+\varepsilon_t+\gamma_t}
	\ge  \One{E_{t}} r_t(a) - \frac{\varepsilon_t + \gamma_t}{\pib(a)},
\end{align*}
where in the first inequality we exploited that $\tr_{t}(a)$ is nonnegative, in the second one we used that $E_{t}$
is $\F_{t-1}$-measurable and the defining property of the good event, and in the last one we simplified some
expressions. Thus, we have
\begin{align*}
	\EE{\sumt \suma \pi^*(a) r_t(a)} & \le \EE{\sumt \suma \pi^*(a) \pa{\tr_t(a) + (1-\One{E_t})r_{t}(a) +
			\frac{\varepsilon_t + \gamma_t}{\pib(a)}}}
	\\
	                                 & = \EE{\sumt \suma \pi^*(a) \tr_t(a)} + \sumt \delta_t + \suma 
\frac{\pi^*(a)}{\pib(a)}\sumt\pa{2\varepsilon_t +
		\frac{\eta}{2}},
\end{align*}
where in the last line we recalled that $\gamma_t = \varepsilon_t + \eta/2$.
Putting the two bounds together, we arrive to
\begin{align*}
	\EE{\sumt \suma \pa{\pi^*(a) r_{t}(a) - \pi_t(a) r_t(a)}} & \leq \frac{\log K}{\eta} + \pa{1 + \frac{2}{\eta}}\sumt
	\delta_t  + \pa{ \frac{\eta n}{2} + 2\sumt \varepsilon_t} \suma \frac{\pi^*(a)}{\pi_b(a)}                           
\\
\end{align*}
Finally, we set $\delta_1=0\,,\,\delta_t = (t-1)^{-2}$ so that we have $\sumt \delta_t = \pi^2/6 \le 
2$ and we can write
\begin{align*}
    \sumt \varepsilon_t = 1+ \sum_{t=1}^{\n-1} \sqrt{\frac{\log(Kt^2)}{2t}} \le 2\sqrt{n\log(Kn)},
\end{align*}
where we also used the standard upper bound $\sum_{t=1}^n 1/\sqrt{t} \le 2\sqrt{n}$.
Putting everything together, we finally get
\[
	\EE{\sumt \suma \pa{\pi^*(a) r_{t}(a) - \pi_t(a) r_t(a)}} \le \frac{16 + \log K}{\eta} +
	\pa{\frac{\eta n}{2} + 2\sqrt{n\log(Kn)}} \coverage(\pi^*;\pib) + 2.
\]
Setting $\eta = \sqrt{\frac{\log K}{n}}$ concludes the proof.


\section{The proof of Theorem~\ref{thm:linear}}\label{app:linear}
The proof combines ideas from the previous two proofs with ideas from \citet{Neu2020} to deal with the contextual
aspect of the problem setting.
In the following, let $\iwr_t(x,a)=\siprod{\htheta_t}{\varphi(x,a)}$.
As a starting point, we fix a context $x\in\X$ and define the estimated regret in
context $x$ against comparator $\pi^*$ as
\begin{align}
	\wh{\Reg}(\pi^*,x) = \sumt \suma \left(\pi^*(a \mid x) - \pi_t(a \mid x)\right) \iwr_t(x, a).
\end{align}
The following lemma gives a bound on the above quantity:
\begin{lemma}\label{lem:prod_classic_bound}
	Suppose that $\eta \hr_t(x,a) \ge -1/2$ holds for all $x,a$. Then, for any fixed $x$ and $\pi^*$,
	\begin{equation}\label{eq:prod_classic_bound}
		\wh{\Reg}(\pi^*, x) \leq \frac{\log K}{\eta} + \eta\sumt \suma
		\pi^*(a\mid x) \pa{\wh{r}_t(x,a)}^2.
	\end{equation}
\end{lemma}
The proof follows from a careful combination of techniques by \citet{CBMS07} and \citet{Neu2020}, and is deferred to
Appendix~\ref{app:prod_proof}. We proceed by noting that for any fixed $x$, the second term in the bound can be bounded
as follows:
\begin{equation}
	\begin{split}
		\EEt{ \pa{\wh{r}_t(x,a)}^2}
		&= \EEt{\pa{\rb_t}^2\varphi(x,a)\transpose \Vinv \varphi(X_t,\ab_t)  \varphi(X_t,\ab_t)^\top \Vinv
			\varphi(x,a)} \\
		&\le \varphi(x,a)^\top \Vinv \V(\pi_B) \Vinv \varphi(x,a) = \varphi(x,a)\transpose\Vinv \varphi(x,a) \\
		&= \trace{\Vinv \varphi(x,a)\varphi(x,a)\transpose},
	\end{split}
\end{equation}
where we have used $\rb_t\le 1$ in the inequality.
Furthermore, in order to use the lemma, we first need to verify that its precondition is satisfied. To this end, notice
that
\begin{align*}
	\left|\iwr_t(x,a)\right| & = \left|R_t\phi(x,a)\transpose \pa{\V(\pib)}^{-1}\phi(X_t,A_t^b)\right|
	%
	\leq \frac{\sup_{x,a} \left\|\phi(x,a)\right\|^2_2}{\lambda_{\min}(\V(\pib))},
\end{align*}
which follows from a straightforward applicaiton of the Cauchy--Schwarz inequality.
Thus, the condition on $\eta$  we
impose in the theorem guarantees that $\eta |\hr_t(x,a)|\le 1/2$.
Now we are in position to invoke Lemma~\ref{lem:prod_classic_bound}, albeit with a specific choice for the context $x$.
Specifically, we let $X_0$ be a ``ghost sample'' drawn independently from the context distribution for the sake
analysis, and apply Lemma~\ref{lem:prod_classic_bound} to obtain
\begin{equation}\label{eq:regret_X0}
	\wh{\Reg}(\pi^*, X_0) \leq \frac{\log K}{\eta} + \eta\sumt \suma \pi^*(a\mid X_0) \trace{\Vinv
		\varphi(X_0,a)\varphi(X_0,a)\transpose}.
\end{equation}
Then, a straightforward calculation inspired by the analysis of \citet{Neu2020} shows that the left-hand side is
related to the expected regret as
\begin{equation}\label{eq:regret_decomposition}
	\EE{\wh{\Reg}(\pi^*, X_0)} = \EE{\sumt\suma\pa{\pi^*(a|X_t) - \pi_t(a|X_t)} r_t(X_t,a)}.
\end{equation}
For completeness, we include this calculation in Appendix~\ref{app:regret_calculation}. The same technique can be used
to deal with the term on the right-hand side as follows:
\begin{align*}
	 & \EE{\sumt \suma \pi^*(a\mid X_0) \Tr\Big( {\Vinv \phi(X_0,a)\phi(X_0,a)\transpose} \Big)}   \\
	 & = \EE{\sumt \Tr{\Big( \Vinv \suma \pi^*(a\mid X_0) \phi(X_0,a)\phi(X_0,a)\transpose} \Big)} \\
	 & = \EE{\sumt  \Tr{\Big( \Vinv \V(\pi^*) \Big)}} = \ccoverage(\pi^*;\pib).
\end{align*}
Thus, taking expectations of both sides of Equation~\eqref{eq:regret_X0} and using the above two results concludes the
proof.
\qed
	
\subsection{The proof of the regret decomposition of 
Equation~\eqref{eq:regret_decomposition}}\label{app:regret_calculation}
We start by fixing an arbitrary $x$ and defining the following notion of pseudo-regret in context $x$:
\[
 \Reg(\pi^*, x) = \sumt \suma \left(\pi^*(a \mid x) - \pi_t(a \mid x)\right) r_t(x,a).
\]
We first note that $\E{\wh{\Reg}(\pi^*,x)}=\E{\Reg(\pi^*,x)}$ holds
    thanks to the unbiasedness of $\iwr_t$ and the independence of $\pi_t$ and
    $\iwr_t$. In particular, this follows from the following derivation:
    \begin{align*}
        \E{\wh{\Reg}(\pi^*, x)} &= \E*{\sumt \Es[\Big]{t}{\suma \left(\pi^*(a \mid x) - \pi_t(a \mid x)\right) 
\iwr_t(x, a)}} \\
        &= \E*{\sumt \suma \left(\pi^*(a \mid x) - \pi_t(a \mid x)\right) \Es[\Big]{t}{\iwr_t(x, a)}} \\
        &= \E*{\sumt \suma \left(\pi^*(a \mid x) - \pi_t(a \mid x)\right) r_t(x, a)} = \E{\Reg(\pi^*, x)}, \\
    \end{align*}
    where we used the tower rule of expectation in the first step, the fact that $\pi_t$ is $\F_{t-1}$-measurable in 
the second step, and the unbiasedness of the reward estimator in the last step.
    To relate $\E{\Reg(\pi^*, x)}$ and the true expected Regret $\E{\Reg(\pi^*)}$, we consider the random variable 
$\Reg(\pi^*,X_0)$ with $X_0$ being a ghost sample drawn from the context distribution 
    independently from the history of contexts $(X_t)_{t=1}^n$. Then, we can write the expectation of this random 
variable as
    \begin{align*}
        & \E{\Reg(\pi^*, X_0)} =
        \E*{\sumt \suma \left(\pi^*(a \mid X_0) - \pi_t(a \mid X_0)\right) r_t(X_0, a)} \\
        &= \E*{\sumt \Es[\Big]{t}{\suma \left(\pi^*(a \mid X_0) - \pi_t(a \mid X_0)\right) r_t(X_0, a)}} \\
        &= \E*{\sumt \Es[\Big]{t}{\suma \left(\pi^*(a \mid X_t) - \pi_t(a \mid X_t)\right) r_t(X_t, a)}} = 
\E{\Reg(\pi^*)},
    \end{align*}
    where the second line uses the tower rule of expectation and the third one the fact that $X_0$ is distributed 
identically with $X_t$ given $\F_{t-1}$. This concludes the proof.
\qed

\subsection{Proof of Lemma~\ref{lem:prod_classic_bound}}\label{app:prod_proof}
The proof is inspired by the classic \prodd analysis of \citet{CBMS07}, and follows from similar arguments as the proof 
of Lemma~\ref{lem:rbound_partial}. The main adjustment we need to these proofs is that now we have to include contexts 
in our derivations. To this end, let us fix one context $x\in\X$ and suppose that the condition of the theorem is 
satisfied: $\eta\,\iwr_t(x,a)\geq -1/2$ for all actions $a\in\A$.

As before, we will study the evolution of the potential function $\frac{1}{\eta} \log \frac{W_{n+1}(x)}{W_1(x)}$.
For every action $\bar{a}\in\A$ we have:
\begin{align*}
    \frac{1}{\eta} \log W_{n+1}(x) &= \frac{1}{\eta} \log \suma \prod_{t=1}^\n (1+\eta \iwr_t(x, a)) 
    \geq \frac{1}{\eta} \log \prod_{t=1}^\n (1+\eta \iwr_t(x, \bar{a}))
    \\
    & = \frac{1}{\eta} \sumt \log (1+\eta \iwr_t(x, \bar{a}))
    \geq \sumt \big( \iwr_t(x, \bar{a}) -  \eta\pa{\iwr_t(x, \bar{a})}^2 \big),
\end{align*}
where we used our condition on the magnitude of the reward estimates twice: once to use $(1-\eta \hr_t(x,a)) \ge 0$ in 
the first line and once when using the elementary inequality $\log(1+z) \geq z - z^2$ that holds for all $z\geq 
-1/2$ in the second line.
%
Moreover, we can upper-bound the potential as 
\begin{align*}
    \frac{1}{\eta}\log W_{n+1}(x) &= \frac{1}{\eta}\log W_1 + \frac{1}{\eta}\log \prod_{t=1}^n \frac{W_{t+1}(x)}{W_t(x)}
    = \frac{\log K}{\eta} + \frac{1}{\eta}\sumt \log \frac{W_{t+1}(x)}{W_t(x)} \\
    &= \frac{\log K}{\eta} + \frac{1}{\eta}\sumt \log \suma \frac{ w_t(x,a)}{W_t(x)}(1+\eta\,\iwr_t(x,a))
    &\big(\text{def. of $W_{t+1}$}\big) \\
    &= \frac{\log K}{\eta} + \frac{1}{\eta}\sumt \log \suma \pi_t(a\mid x)(1+\eta\,\iwr_t(x,a))
    &\big(\text{def. of $\pi_{t}$}\big) \\
    &= \frac{\log K}{\eta} + \frac{1}{\eta}\sumt \log \Big( 1 + \eta \suma \pi_t(a\mid x)\iwr_t(x,a) \Big) \\
    &\leq \frac{\log K}{\eta}+ \sumt \suma \pi_t(a\mid x)\iwr_t(x,a),
\end{align*}
where we used the inequality $\log(1+z) \le z$ that holds for all $z > -1$.

Combining the lower bound and upper bound, we obtain
$$
\sumt \Big( \iwr_t(x, \bar{a}) - \suma \pi_t(a\mid x)\iwr_t(x, a)\Big) \leq \frac{\log K}{\eta} + 
\eta\sumt\pa{\iwr_t(x,\bar{a})}^2.
$$

Multiplying both sides by $\pi^*(\bar{a}\mid x)$ and summing over all actions
$\bar{a}\in\A$ yields the desired result.
\qed

\end{document}